\documentclass{article}

\usepackage{microtype}
\usepackage{graphicx}
\usepackage{subfigure}
\usepackage{booktabs} 

\usepackage{hyperref}

\usepackage[accepted]{icml2023}

\usepackage{amsmath}
\usepackage{amssymb}
\usepackage{mathtools}
\usepackage{amsthm}

\usepackage[capitalize,noabbrev]{cleveref}

\theoremstyle{plain}
\newtheorem{theorem}{Theorem}[section]

\theoremstyle{definition}
\newtheorem{definition}[theorem]{Definition}

\theoremstyle{remark}

\usepackage[textsize=tiny]{todonotes}

\icmltitlerunning{Providing Recourse While Minimizing Training Data Leakage}

\begin{document}

\twocolumn[
\icmltitle{Accurate, Explainable, and Private Models: \\Providing Recourse While Minimizing Training Data Leakage}



\icmlsetsymbol{equal}{*}

\begin{icmlauthorlist}
\icmlauthor{Catherine Huang}{equal,harvcs}
\icmlauthor{Chelse Swoopes}{equal,harvcs}
\icmlauthor{Christina Xiao}{equal,harvcs}
\icmlauthor{Jiaqi Ma}{harvcs}
\icmlauthor{Hima Lakkaraju}{harvcs}
\end{icmlauthorlist}

\icmlaffiliation{harvcs}{Department of Computer Science, Harvard University School of Engineering and Applied Sciences, Cambridge, Massachusetts}

\icmlcorrespondingauthor{Catherine Huang}{catherinehuang@college.harvard.edu}
\icmlcorrespondingauthor{Chelse Swoopes}{cswoopes@g.harvard.edu}
\icmlcorrespondingauthor{Christina Xiao}{christinaxiao@college.harvard.edu}
\icmlcorrespondingauthor{Jiaqi Ma}{jima@hbs.edu
}
\icmlcorrespondingauthor{Hima Lakkaraju}{hlakkaraju@hbs.edu}

\icmlkeywords{algorithmic recourse, adversarial machine learning, privacy, explainability, interpretability}

\vskip 0.3in
]



\printAffiliationsAndNotice{\icmlEqualContribution} 

\begin{abstract}



Machine learning models are increasingly utilized across impactful domains to predict individual outcomes. As such, many models provide algorithmic recourse to individuals who receive negative outcomes. However, recourse can be leveraged by adversaries to disclose private information. This work presents the first attempt at mitigating such attacks. We present two novel methods to generate differentially private recourse: Differentially Private Model (\texttt{DPM}) and Laplace Recourse (\texttt{LR}). Using logistic regression classifiers and real world and synthetic datasets, we find that \texttt{DPM} and \texttt{LR} perform well in reducing what an adversary can infer, especially at low \texttt{FPR}. When training dataset size is large enough, we find particular success in preventing privacy leakage while maintaining model and recourse accuracy with our novel \texttt{LR} method.
\end{abstract}

\section{Introduction} 
\label{section:introduction}
\noindent Explainability and privacy are two important pillars of trustworthy machine learning (ML), but they are often viewed as conflicting. A right to privacy is often viewed as a limit on transparency \cite{transparency}. Still, users may want an explanation of how a ML system works or why it gave a particular outcome, even as they want their personal data to be kept private \cite{transparency} \cite{featureattack}. As users are increasingly impacted by negative model predictions in domains such as healthcare, medicine, and criminal justice, there is a growing emphasis on providing algorithmic recourse to individuals, so that they can understand and contest decisions, or alter their behavior to achieve a preferred outcome \cite{wachter2017counterfactual}. 

Recourse commonly takes the form of counterfactual explanations (CFEs), which highlight what feature changes a user would need to make for a model's predicted label to change \cite{recourseattack}. However, recent work by Pawelczyk et al. showed severe privacy risks with CFEs, as they developed two successful membership inference (MI) attacks utilizing CFEs to leak training data \yrcite{recourseattack}. 

In this work, we develop private recourse methods that protect against MI attacks while maintaining reasonable model accuracy. We hypothesize that the mathematical framework \emph{differential privacy (DP)} can be used to create these private recourses \cite{dwork}. Investigating the privacy, accuracy, and explainability trade-off is an under-explored area. Our work is the first attempt at mitigating the MI attacks presented by Pawelczyk et al. and the first work generating recourses in a DP manner.

\section{Related Work}
The research area of machine learning explainability and privacy is quite recent, and thus contains a smaller amount of prior work. Furthermore, most works have revolved around attacks leveraging explainability to the detriment of privacy, rather than creating explainable and still private models. Shokri et al. demonstrated that feature-based explanations may leak sensitive information about training data \yrcite{featureattack}. Milli et al. demonstrated that gradient-based explanations could be used to quickly reconstruct the underlying model \yrcite{milli2019model}. A\"{i}vodji et al. showed that CFEs could also be leveraged for highly accurate model extraction attacks \yrcite{aivodji2020model}.

We are interested in CFEs that may expose information on training data. Pawelczyk et al. provided the first work in this area, with two novel counterfactual distance-based MI attacks: \textbf{1)} thresholding on counterfactual distance (\texttt{CFD}), and \textbf{2)} likelihood ratio test using counterfactual distance (\texttt{CFD LRT}) \yrcite{recourseattack}. We discuss the details of these attacks, which are the basis of our work, in Section \ref{section:preliminaries}. With enough recourse queries, adversaries can use these attacks to reconstruct the training data of a non-private, recourse-supporting model \cite{recourseattack}.

\section{Preliminaries}
\label{section:preliminaries}
\subsection{Algorithmic Recourse}

We base our recourse definition off of Wachter et al.'s first work in this area \yrcite{wachter2017counterfactual}. Wachter et al.'s definition is also adopted by our motivating work \cite{recourseattack} and other explainability-privacy papers. 
\begin{definition}

Let $x \in \mathcal{X}$ be a data observation that received a negative label when fed through $f_{\theta}$, where $f_{\theta}: \mathcal{X} \rightarrow \mathcal{Y}$ is a classifier model parameterized by $\theta$, $\mathcal{X} \in \mathbb{R}^d$, and $\mathcal{Y} \in \{0, 1\}$. 

Finding an \textbf{algorithmic recourse} for $x$ means finding a counterfactual \[x' = x + \delta: f_{\theta}(x') = f(x + \delta) = 1.\] We aim to minimize the cost $c(x, x')$ to change $x$ to $x'$ so that the recourse is easily implementable; in practice, $\ell_2$ distance is commonly used as a cost function.
\end{definition}

\subsection{Counterfactual Distance-Based Membership Inference Attacks for ML Models} \label{section:miattacks}
MI attacks infer whether an instance $x$ belongs to the training data for $f_{\theta}$. Across the literature, MI attacks are commonly \textit{loss-based}, following the intuition that models have lower loss on instances observed during training \cite{yeom} \cite{carlini}. Pawelczyk et al.'s novel counterfactual distance-based MI attacks show that counterfactual distance can also be used to leak training data \yrcite{recourseattack}.

\textbf{Thresholding on counterfactual distance (\texttt{CFD}):} Intuitively, during training, the decision boundary is pushed away from training points (as in margin maximization), resulting in test set points being closer to the decision boundary.
\[M_{\texttt{CFD}}(x) = \begin{cases}
    \texttt{MEMBER}, & \text{if } c(x, x') \ge \tau_{\mathcal{D}} \\
    \texttt{NON-MEMBER}, & \text{if } c(x, x') < \tau_{\mathcal{D}}
  \end{cases}. \]

\textbf{Counterfactual distance likelihood ratio test (\texttt{CFD LRT}):} In this attack, the adversary trains shadow models to estimate the likelihood ratio \[\Lambda = \displaystyle\frac{\Pr[c(x, x') | x \in \mathcal{D}_t]}{\Pr[c(x, x') | x \not\in \mathcal{D}_t]}.\] The attack then thresholds on this $\Lambda$. 

See Appendix \ref{appendix:cfdlrt} for the formulation and implementation details of this attack.

\subsection{Differential Privacy} \label{section:dpdefinition}

Our solution involves using \emph{differential privacy} (DP) to counteract the success of the counterfactual distance-based MI attacks in Section \ref{section:miattacks}. DP is a mathematically provable definition of privacy that provides a quantifiable metric of privacy loss, providing a computational method whose output is random enough to obscure any single participant's presence in the training data \cite{dwork}. 

\begin{definition} A randomized mechanism $\mathcal{M}$ with domain $\mathcal{D}$ and range $\mathcal{R}$ satisfies \textbf{$\varepsilon$-differential privacy} ($\varepsilon$-DP) if for any two adjacent input datasets $d, d'\in \mathcal{D}$ differing by one row, and any subset of outputs $S \subseteq \mathcal{R}$ \[Pr[\mathcal{M}(d) \in S] \le e^{\varepsilon}Pr[\mathcal{M}(d') \in S].\] The $\varepsilon$ parameter represents \textit{privacy loss}: the lower the $\varepsilon$, the stronger the privacy protection.

\end{definition}

\subsubsection{Laplace Mechanism of Differential Privacy} \label{section:laplacemech}

The Laplace Mechanism, a widely used DP mechanism, is useful on numerical queries. It involves adding Laplacian distributed random noise on the output of a sensitive query \cite{dwork}.

\begin{definition} \label{definition:laplace} For sensitive query $f(d)$ on input dataset $d$, the \textbf{$\varepsilon$-DP Laplace Mechanism} $\mathcal{M}_{Lap}$ is defined as \[\mathcal{M}_{Lap}(d) = f(d) + \mathrm{Laplace}(GS_f/\varepsilon),\] where $\mathrm{Laplace}(GS_f/\varepsilon)$ is a Laplace random variable with scale parameter $GS_f/\varepsilon$. 

$GS_f$ is the \textit{global sensitivity} of query $f$, bounding how much the sensitive query outcome can change across any two possible neighboring datasets $d, d' \in \mathcal{D}$: $$GS_f = \max\limits_{d, d'} || f(d) - f(d')||_1.$$

\end{definition}

\subsubsection{Differential Privacy Under Post-Processing} \label{section:postprocessing}

Dwork et al. prove that once a quantity is ``made private" through DP, it cannot be subsequently "made un-private" \yrcite{dwork}. This is formalized in the following theorem.

\begin{theorem}
If mechanism $M$ is $\varepsilon$-DP, and $G$ is an arbitrary deterministic mapping, then $G \circ M$ is also $\varepsilon$-DP.
\end{theorem}

\section{Problem Statement \& Methodology}
\subsection{Problem Statement} \label{section:problemstatement}

We hypothesize that DP can be used as a privacy preservation mechanism to protect algorithmic recourse models from MI attacks. To thoroughly evaluate the extent and nature of the impact of DP on CFD based attack success, we evaluate the influence of the following changes to our \textit{ML classifier $\rightarrow$ algorithmic recourse $\rightarrow$ MI attack} pipeline:
\begin{itemize}
\setlength\itemsep{0.01em}
    \item The presence versus absence of differential privacy.
    \item The particular DP mechanism being used: \texttt{DPM} versus \texttt{LR} (see Section \ref{section:dprecourse}).
    \item The privacy protection strength $\varepsilon$ of the DP mechanism.
    \item The type of MI attack: \texttt{CFD} versus \texttt{CFD LRT}.
    \item The dataset, whether synthetic or real-world.
    \item The dimensionality of the data, for synthetic data.
\end{itemize}

\subsection{Methodology}

\subsubsection{Recourse for Logistic Regression Classifiers}

Logistic regression, our classifier of choice, has weights $\mathbf{w}$ after training which it uses to output a probability score: $f(x) = \mathbf{w}^T x = \log \frac{\Pr(y = 1 | x)}{1 - \Pr(y=1 | x)}.$ In a linear model such as this one, it is standard for the counterfactual distance of instance $x$ to be calculated using the $\ell_2$ norm from $f(x)$ to the target score $s$ in logistic regression space: \[c(x, x') = \frac{s-f(x)}{||\mathbf{w}||_2^2} \mathbf{w}.\] This is the counterfactual distance calculation method we use in our methodology. In our experiments, we set our decision boundary threshold to $s = 0$, which corresponds to the fitted $\Pr(y = 1| x)$ being equal to $\frac{1}{2}$ at the threshold.

\subsubsection{Differentially Private Recourse Generation Methods} \label{section:dprecourse}

\textbf{Differentially Private Model (\texttt{DPM})}

Our first DP method trains the \textit{underlying} logistic regression classifier with DP. By post-processing of DP (see Section \ref{section:postprocessing}), an $\varepsilon$-DP logistic regression model gives rise to $\varepsilon$-DP counterfactual recourse. We use IBM's \texttt{diffprivlib} library \cite{diffprivlib}, which offers an implementation of DP logistic regression based on Chaudhuri et al.'s formulation of a DP empirical risk minimization mechanism \yrcite{dplogreg}. 

\textbf{Differentially Private Laplace Recourse (\texttt{LR})} 

We propose a novel method for DP post-hoc computation of counterfactual recourse that does not touch the underlying logistic regression model training process. The method is as follows:

\begin{enumerate}
\setlength\itemsep{0.03em}
    \item Apply the Laplace mechanism onto the \textit{predicted probability score} $\Pr'(y=1|x) = \Pr(y=1|x) + \mathrm{Laplace}(1/\varepsilon).$ 
    \item Clamp $\Pr'(y=1|x)$ to $[0, 1]$ so that $\Pr'(y=1|x)$ can still be interpreted as a probability.
    \item Calculate noisy logistic regression score $f'(x) = \log \frac{\Pr'(y=1|x)}{1 - \Pr'(y=1|x)}.$
    \item Calculate noisy CFD: $\mathcal{M}_{\texttt{CFD}, Lap}(x) = \frac{s-f'(x)}{||\mathbf{w}||_2^2} \mathbf{w}.$
\end{enumerate}

\textbf{Claim:} The above method is $\varepsilon$-DP.

\textbf{Explanation:} Our method begins with applying the Laplace mechanism (see Section \ref{section:laplacemech}) on the predicted probability $\Pr(y=1|x)$, for data instance $x$. We claim that $GS_{P(y=1|x)} = 1$, i.e. the vector of all ones. First, note that $\Pr(y=1|x)$ is a probability vector $\in [0, 1]^d$. For any two possible datasets $d, d' \in \mathcal{D}$, if we calculate $\Pr(y=1|x)_d$ using the classifier trained on $d$, and then calculate $\Pr(y=1|x)_{d'}$ using the classifier trained on $d'$, the two probabilities can differ in $\ell_1$ distance by at most 1. Steps 2-4 in the method are post-processing functions applied to $\Pr'(y=1|x)$. By post-processing of DP (see Section \ref{section:postprocessing}), we retain $\varepsilon$-DP.

\section{Experimental Results}
\subsection{Setup}

\textbf{Datasets:} To stay relevant and methodologically consistent with the motivating work \cite{recourseattack}, we use similar datasets. For real world data, we use the datasets \textbf{1)} Heloc (Home Equity Line of Credit) \cite{heloc} ($d=23$), which scores whether individuals will repay their Heloc accounts within a fixed time window, \textbf{2)} MNIST \cite{mnist}, which contains $28 \times 28$ pixel gray-scale images of handwritten digits between 0 and 9, and \textbf{3)} Adult \cite{adult} ($d=14$), a variant of the 1994 Census database that labels whether an individual has annual income greater than \$50,000. However, based on the minimal success of baseline \texttt{CFD} attacks on Adult, we omit this analysis from this writeup, although our results are presented in the Appendix \ref{section:adultappendix}.

For synthetic data, we follow Pawelczyk et al. \yrcite{recourseattack} and Shokri et al \yrcite{featureattack}: For $d \in \{100, 1000, 5000, 7000\}$, we randomly choose a vertex from a $d$-dimensional hypercube and sample $n=5000$ random variables from a Gaussian distribution centered at the vertex with unit variance.

\textbf{Pre-processing:} We use 5000 data entries for each model training set, and give the adversary 5000 entries to train their own shadow models. Before model fitting, we pre-process data by removing multicollinear features (with correlation over 0.95), standardizing, and normalizing so that each feature's $\ell_2$ norm is 1.

\textbf{Attack specifications:} For both attacks, we calculate counterfactual distance based on $\ell_2$ norm to the decision boundary. For \texttt{CFD LRT} attacks, we consider two versions with global and local variance estimators, respectively, following the paper that introduced the \texttt{LRT} MI attacks \cite{carlini} that the motivating work \cite{recourseattack} is based upon. In each \texttt{CFD LRT} attack, we train 5 shadow models and 20 ensemble models. 

\textbf{Settings:} We evaluate the baseline setting as proposed by Pawelczyk et al. \yrcite{recourseattack}, alongside our novel DP model (\texttt{DPM}) and Laplace recourse (\texttt{LR}) methods (refer to Section \ref{section:dprecourse}), each with $\varepsilon = 0.5, 1.0$. 

\subsection{Metrics} \label{section:metrics}

Following the motivating work \cite{recourseattack}, we use log-scaled \texttt{ROC} curves (receiver operating characteristic), \texttt{AUC} (area under the curve), and \texttt{BA} (balanced accuracy) to determine the efficacy of the MI attacks. Figs \ref{fig:heloc}, \ref{fig:mnist}, and \ref{fig:synth} show these metrics. Based on motivating literature \cite{carlini}  \cite{recourseattack}, we pay particular attention to successes at low \texttt{FPR}, as a MI attack is still successful if it can identify even a very small subset of the training data with high confidence. 

The DP literature acknowledges the tradeoff of privacy and accuracy \cite{dwork}. Because we aim to create models that are at once accurate, private, and explainable, we also consider the train accuracy on the last ensemble model, and the test accuracy across all 20 ensemble models. To examine recourse accuracy, we also compare the train and test distributions of CFDs and CFD LRTs on synthetic data, under all three settings (baseline, \texttt{LR}, \texttt{DPM}); these results are in Fig \ref{fig:hists}.

\subsection{Results and Analysis}
\label{section:results}

\begin{figure}[h!]
\centering
  \includegraphics[width=0.96\linewidth]{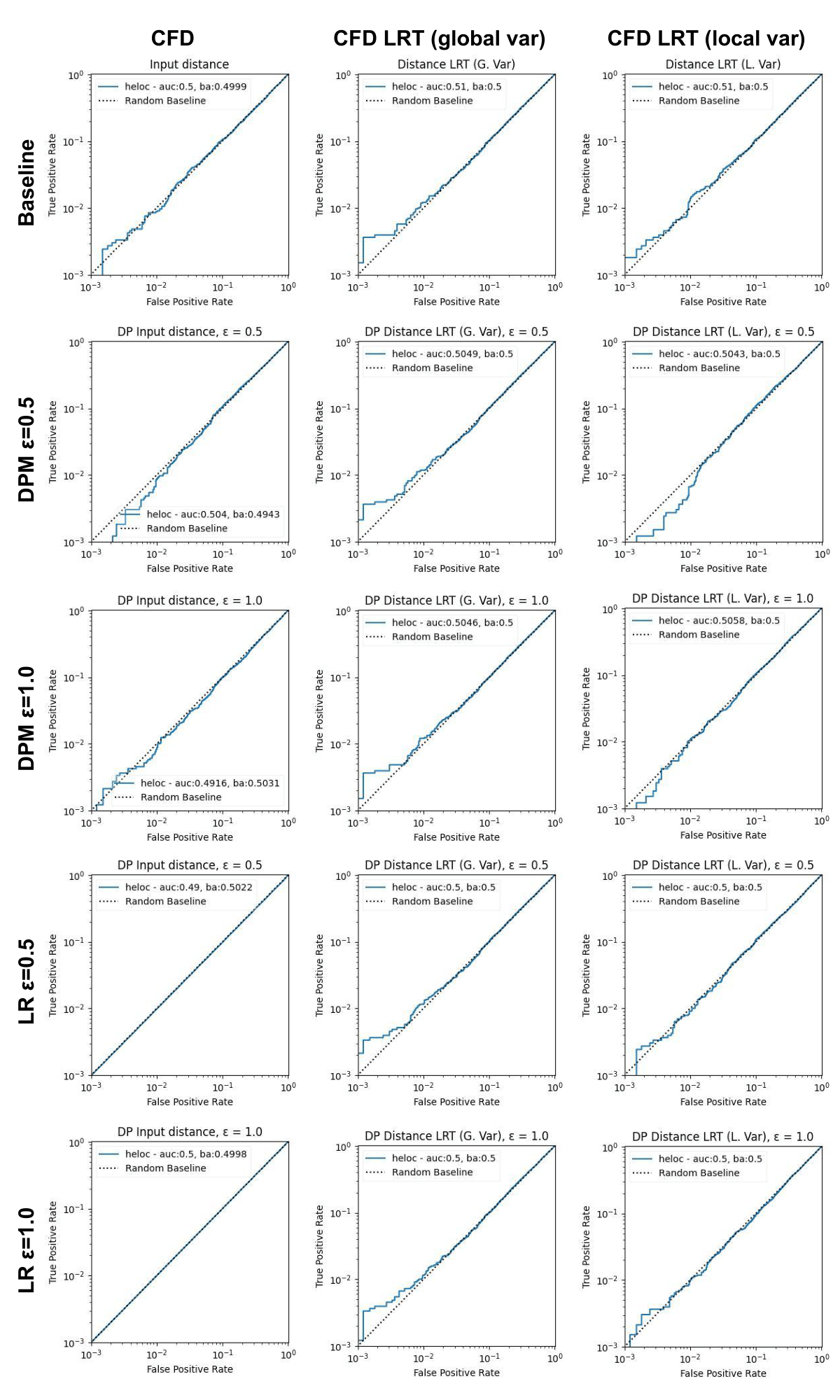}
  \caption{Log-scaled \texttt{ROC} curves (\texttt{TPR} v. \texttt{FPR}), \texttt{AUC}, and \texttt{BA} for all attacks on all settings using Heloc dataset. See \textbf{Privacy:} for analysis.}
  \label{fig:heloc}
  \end{figure}

\begin{figure}[h!]
  \centering
  \includegraphics[width=0.96\linewidth]{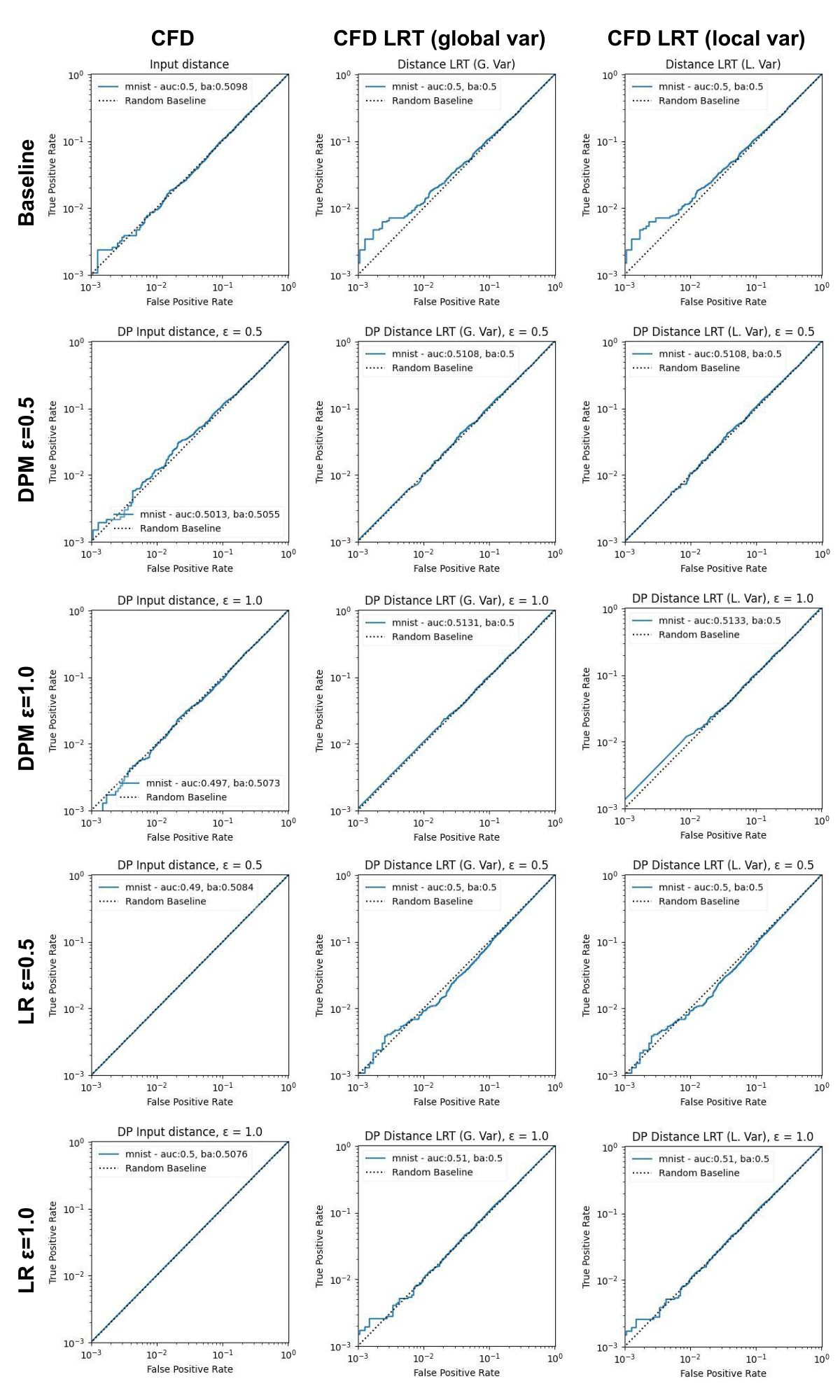}
  \caption{Same as Fig. \ref{fig:heloc}, but using MNIST dataset.}
  \label{fig:mnist}
  \end{figure}

\begin{figure}[h!]
  \centering
  \includegraphics[width=0.96\linewidth]{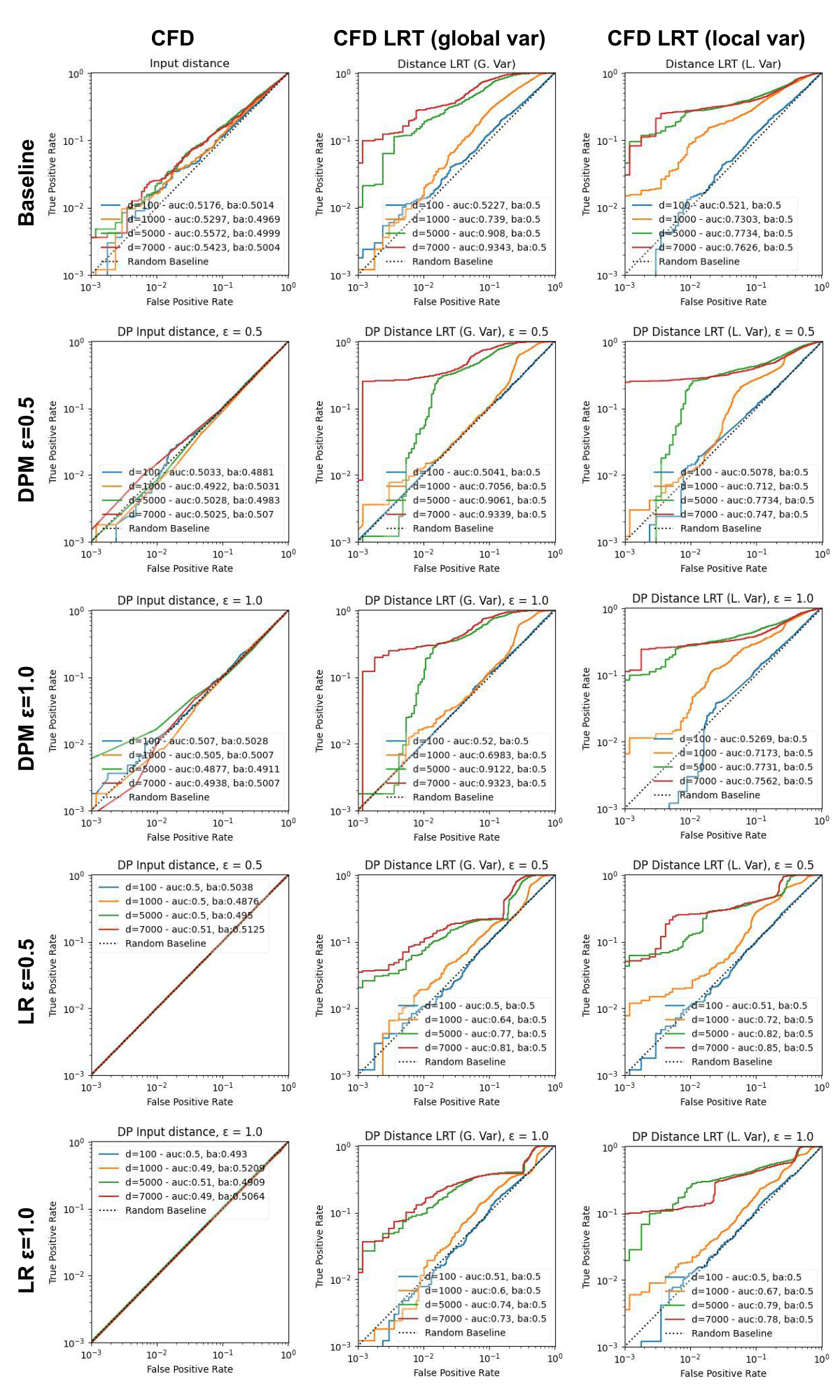}
  \caption{Same as Fig. \ref{fig:heloc}, but using synthetic dataset.}
  \label{fig:synth}
\end{figure}

\begin{figure}[h!]
  \centering
  \includegraphics[width=0.96\linewidth]{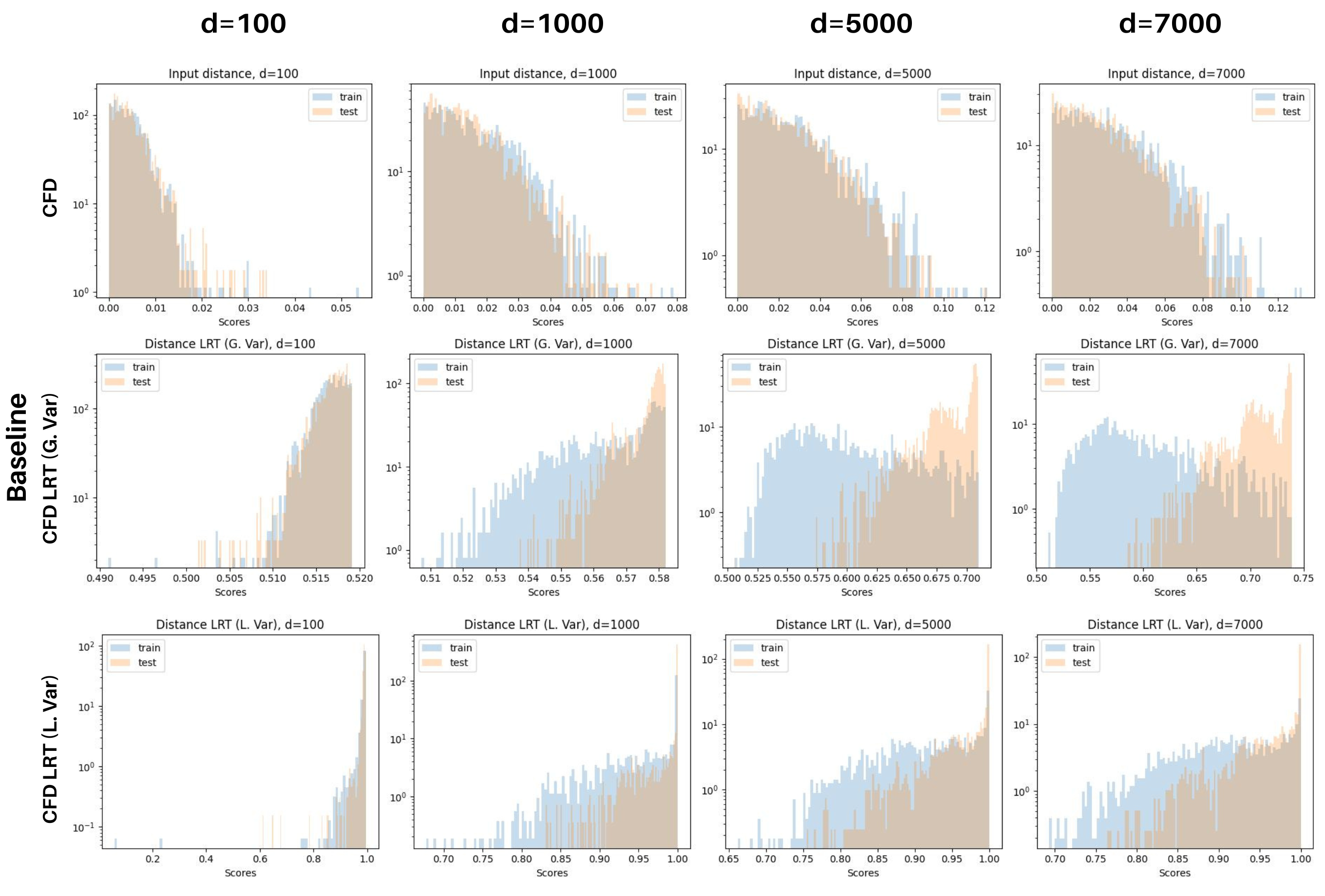}

  \includegraphics[width=0.96\linewidth]{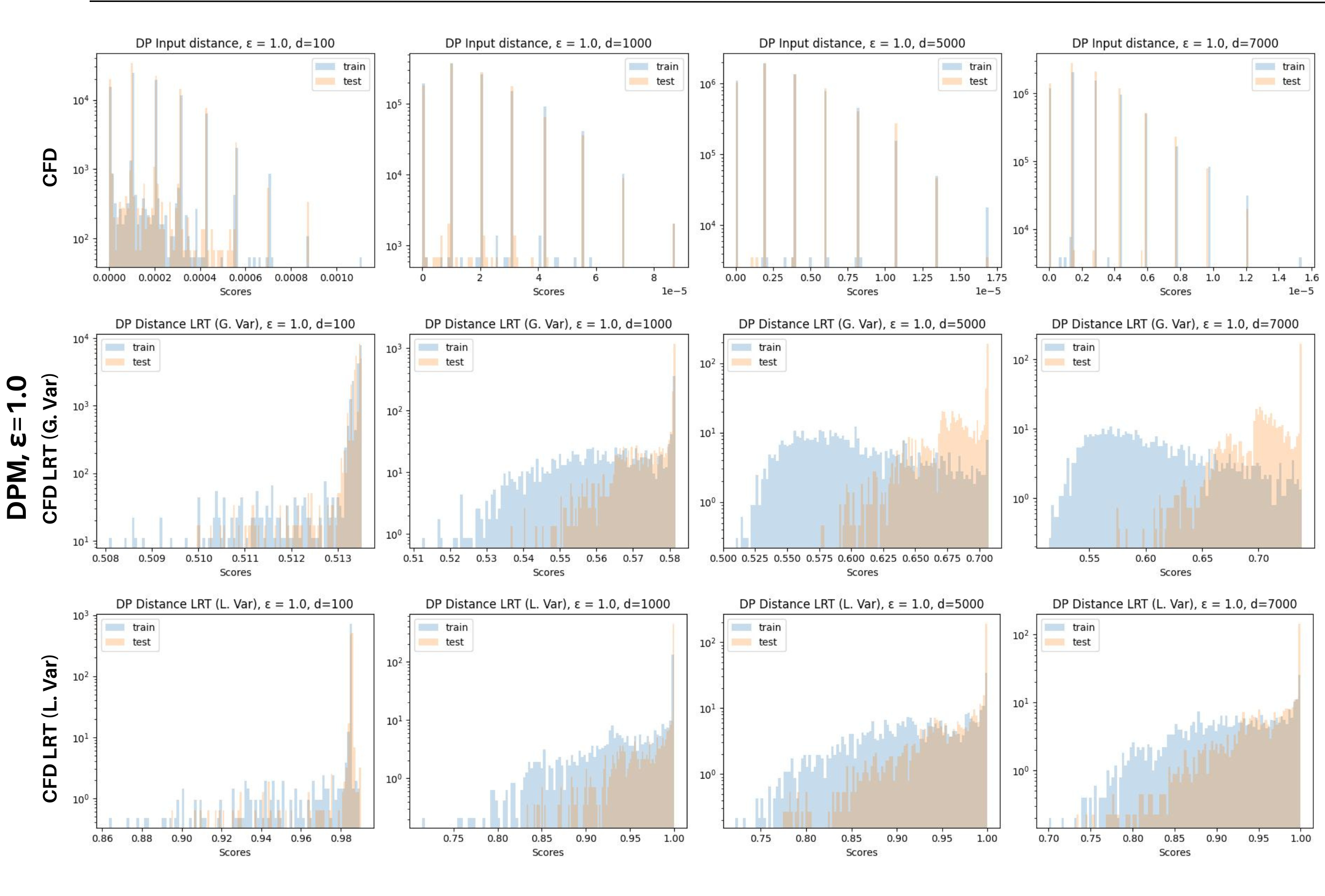}

  \includegraphics[width=0.96\linewidth]{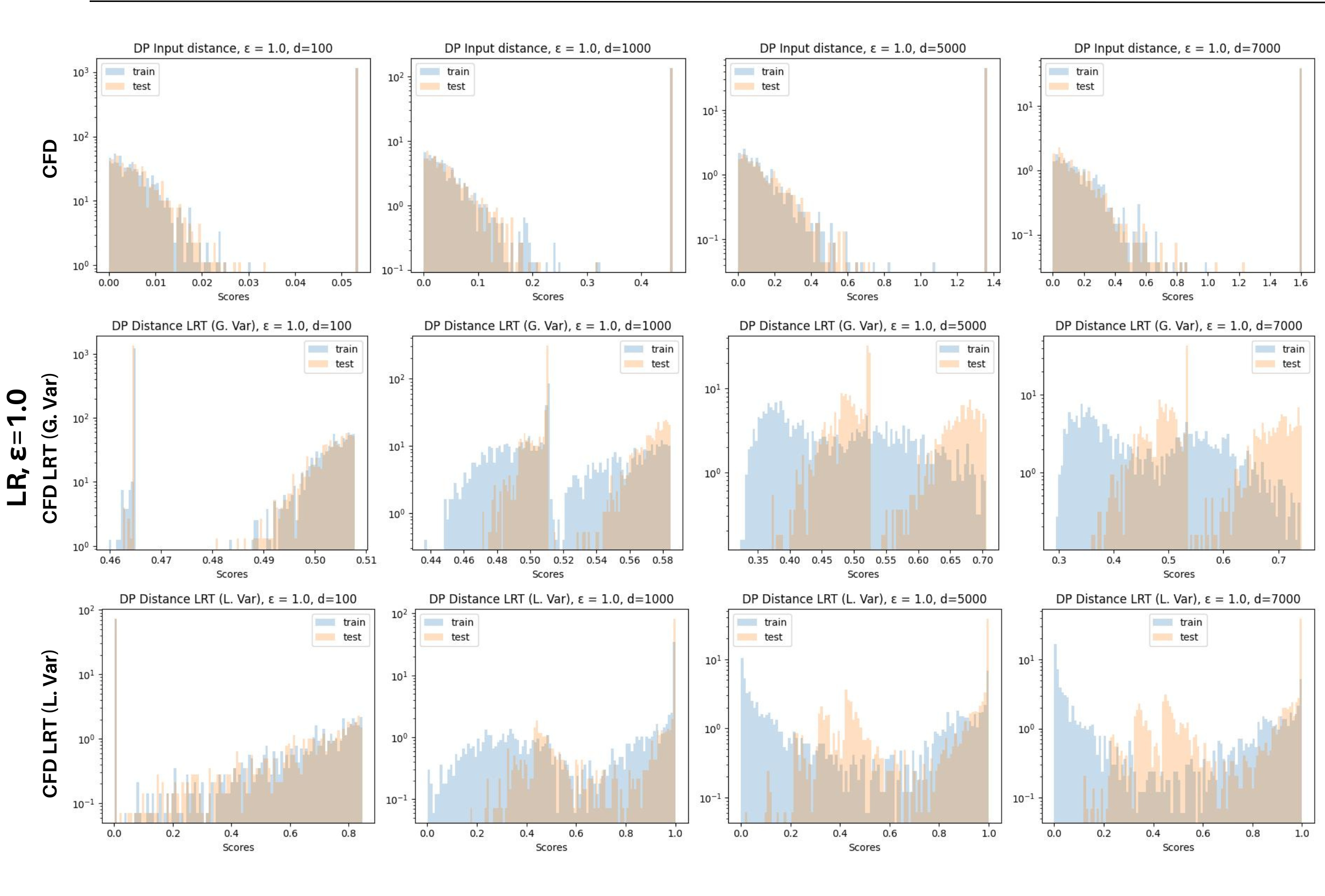}
  
  \caption{Histograms of train and test counterfactual distances (for the Input Distance (\texttt{CFD}) attack) and CFD LRTs (for the LRT attacks), under the synthetic data, $\varepsilon=1.0$ setting. We see that as data dimension increases, train and test distributions differ, even under differentially private recourse generation. This difference is especially pronounced in CFD LRTs. These histograms also offer insight on DP recourse accuracy, explained in Section \ref{section:results}}.
  \label{fig:hists}
\end{figure}

\textbf{Privacy:} In Figs \ref{fig:heloc}, \ref{fig:mnist}, and \ref{fig:synth}, we hoped to see the \texttt{DPM} and \texttt{LR} methods flatten the \texttt{ROC} curves of the baseline, towards the random line with lower \texttt{AUC}, for all attacks---particularly at low \texttt{FPR}, as explained in Section  \ref{section:metrics}. This would show the success of our methods in reducing the efficacy of the MI attacks. Overall, we see that both DP methods generally do flatten the \texttt{ROC} curves from the baseline, particularly at low \texttt{FPR}, across all datasets. For a particularly impressive example, compare \texttt{DPM} $\varepsilon=1.0$ and baseline for \texttt{CFD LRT} (global variance) in Fig. \ref{fig:mnist}. As expected, $\varepsilon=0.5$, with more DP, protects against attacks more successfully than $\varepsilon=1.0$. 

At both values of $\varepsilon$, \texttt{LR} appears highly successful against \texttt{CFD}, with almost exactly random lines and $\texttt{AUC}=0.5$, across all datasets. Similarly, \texttt{DPM} appears slightly more successful against both \texttt{CFD LRT} attacks, particularly in Fig. \ref{fig:mnist}. This makes intuitive sense since both \texttt{LR} and \texttt{CFD} happen on the level of recourse, while \texttt{DPM} and \texttt{CFD LRT} happen on the level of model training. 

The motivating work proposes this theorem: For a $\varepsilon$-DP recourse mechanism—such as our \texttt{LR}—the \texttt{BA} of all attacks is bounded by $\frac{1}{2} + \frac{1-e^{-\varepsilon}}{2}$ \cite{recourseattack}. For $\varepsilon=0.5, 1.0$, this bound is $0.697, 0.816$, respectively. We are pleased to announce that our empirical \texttt{BA} far surpasses this theoretical bound, with \texttt{BA} across nearly all DP methods, attacks, and datasets at approximately 0.5.

\begin{table}[h!]
 \centering
 \caption{Train accuracy of last ensemble model and test accuracy over all 20 ensemble models, using real world datasets. (\texttt{LR} is excluded because it has no impact on model training; we observed very similar empirical results, with only minor variation based on the split of training data points between ensemble models.) We see that training with DP significantly lowers train and test accuracy for both datasets. As expected, $\varepsilon=1.0$ is much more accurate than $\varepsilon=0.5$ in training — but it is only slightly more accurate in testing.}
 \scalebox{0.95}{
 \begin{tabular}{l|rr|rr}
    & Heloc & & Mnist & \\
    & \textbf{Train} & \textbf{Test} & \textbf{Train} & \textbf{Test} \\\hline
  \textbf{Baseline} & 0.9046 & 0.8506 & 1.0 & 0.9393 \\
  \textbf{DPM, $\varepsilon=0.5$} & 0.6079 & 0.5254 & 0.6239 & 0.4795 \\
  \textbf{DPM, $\varepsilon=1.0$} & 0.8011 & 0.5648 & 0.7317 & 0.4985 \\
 \end{tabular}
 }
 \label{tab:realacc}
\end{table}

\begin{table}[h!]
 \centering
 \caption{Same as Tab. \ref{tab:realacc}, but using synthetic datasets. Again, we see that training with DP lowers train and test accuracy across all datasets, and that $\varepsilon=1.0$ provides more accuracy than $\varepsilon=0.5$. However, given the low test accuracy on synthetic data under even the baseline condition, the discrepancy in test accuracy under DP is less noticeable.}
  \scalebox{0.95}{
 \begin{tabular}{l|rr|rr}
    & $d=100$ & & $d=1000$ & \\
    & \textbf{Train} & \textbf{Test} & \textbf{Train} & \textbf{Test} \\\hline
  \textbf{Baseline} & 1.0 & 0.6799 & 1.0 & 0.5619 \\
  \textbf{DPM, $\varepsilon=0.5$} & 0.5323 & 0.4933 & 0.4913 & 0.5016 \\
  \textbf{DPM, $\varepsilon=1.0$} & 0.5386 & 0.5047 & 0.5165 & 0.5029 \\
 \end{tabular}
 }
 \vspace{0.05in}

 \scalebox{0.9}{
 \begin{tabular}{l|rr|rr}
    & $d=5000$ & & $d=7000$ & \\
    & \textbf{Train} & \textbf{Test} & \textbf{Train} & \textbf{Test}  \\\hline
  \textbf{Baseline} & 1.0 & 0.5261 & 1.0 & 0.5215 \\
  \textbf{DPM, $\varepsilon=0.5$} & 0.4551 & 0.4953 & 0.515 & 0.5035 \\
  \textbf{DPM, $\varepsilon=1.0$} & 0.4693 & 0.501 & 0.5244 & 0.5012 \\
 \end{tabular}
 }
 \vspace{0.1in}
 \label{tab:synthacc}
\vspace{-4mm}
\end{table}

\textbf{Model Accuracy:} Considering train versus test accuracies in Tables \ref{tab:realacc} and \ref{tab:synthacc}, we are not enthused about the tradeoff in accuracy and privacy under \texttt{DPM}. \texttt{LR} seems the more fruitful method.

\textbf{Recourse Accuracy:} Another way to assess accuracy is to compare the distributions of the DP CFD calculations with the baseline CFD distribution. These distributions are shown in Fig \ref{fig:hists}. For dimension $d \ge 1000$, both \texttt{DPM} and \texttt{LR} have inaccurate CFD distributions, as seen through large disparities in horizontal axis scaling. For $d=100$, however, \texttt{LR} offers a CFD distribution close to that of the baseline, with the correct scaling. We hypothesize that the high bar to the right in the \texttt{LR} case---the main noticeable difference when compared with baseline---is a result of the clamping in the novel \texttt{LR} method, where we clamp the noisy predicted probability score to [0, 1], so that the score can still be interpreted as a probability. Namely, we believe the bar corresponds to observations that received a negative label, and whose noisy predicted probabilities were originally negative and then clamped to 0. We proceed to test this hypothesis by examining the impact of privacy strength ($\varepsilon$) on the severity of this clamping-induced issue. Not only do these results reveal the privacy-accuracy trade-off (prevalent in and consistent with DP literature), but point to \texttt{LR} as promising for maintaining recourse accuracy.

\textbf{On the Relationship Between Privacy Strength and Recourse Accuracy}

\begin{figure}[h!]
    \centering
    \includegraphics[width=0.48\linewidth]{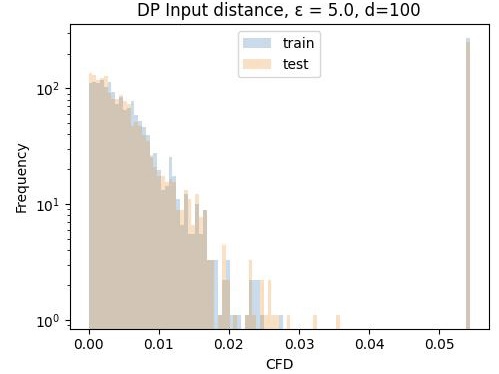} \includegraphics[width=0.48\linewidth]{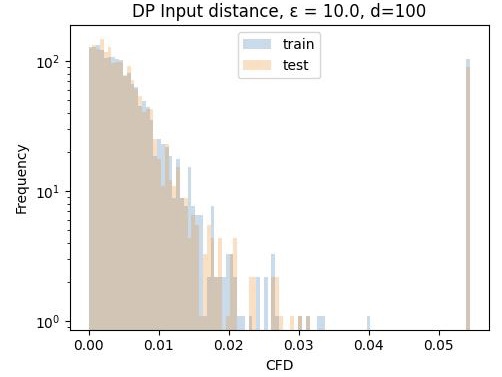} \includegraphics[width=0.48\linewidth]{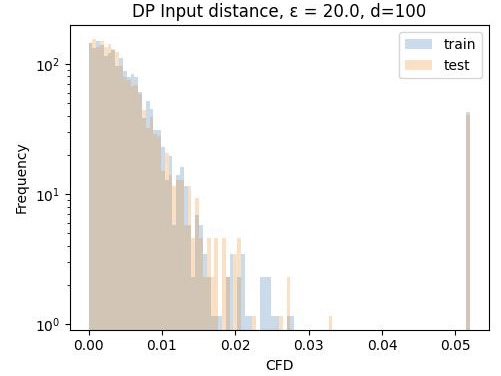}
    \includegraphics[width=0.48\linewidth]{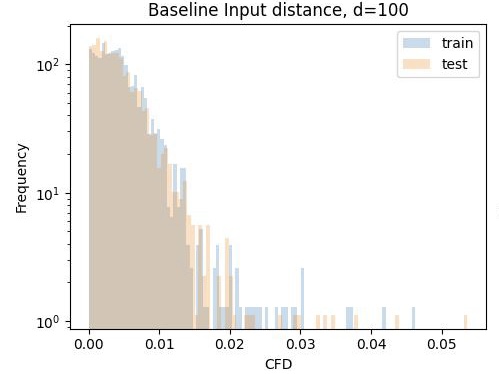} 
    \caption{Histograms of \texttt{LR}-based train and test counterfactual distances from models trained with synthetic data, under the settings $d=100, \varepsilon=\{5, 10, 20\}$. The \texttt{LR}-based CFD distribution approaches the baseline CFD distribution as we increase $\varepsilon$.}
    \label{fig:appendixhists}
\end{figure}

The Laplace Mechanism (Definition \ref{definition:laplace}) involves adding Laplace-distributed random noise to a sensitive numerical query, where the noise's scale parameter is proportional to $1/\varepsilon$. Lower $\varepsilon$ corresponds to stronger privacy protection under DP, but for our \texttt{LR} method, it also means there is higher perturbation of the logistic regression's predicted probability scores, and higher chance that a noisy predicted probability score falls below 0 (upon which it is clamped to 0). The CFD distributions in Figure \ref{fig:hists} illustrate a consequence of this clamping and is a source of accuracy loss of \texttt{LR}-generated recourse. In this appendix, we examine whether decreased privacy protection (higher $\varepsilon$) mitigates this consequence of clamping and yields better CFD distribution accuracy. Figure \ref{fig:appendixhists} shows CFD distributions for $d=100$ synthetic data, under the $\texttt{LR}, \epsilon=\{5, 10, 20\}$ settings.

These findings highlight a privacy-accuracy tradeoff (also referred to as privacy-utility tradeoff in DP literature): the CFD distribution is more accurate for larger values of $\varepsilon$ (i.e. weaker levels of privacy protection).  

Even though the values of $\epsilon$ presented here are larger than those presented in main results (and those used as library default parameters), it is not uncommon for $\varepsilon$ values of up to 10 (and at times up to 20) to be used in the literature: \cite{abadi}, \cite{evaluatingdpinpractice}. Overall, this is a promising result for the accuracy of \texttt{LR}-based recourse.


\section{Future Work}

The accuracy and utility of \texttt{DPM} and \texttt{LR}-generated recourse are worth further exploration; our work is a start. Our recourse accuracy analysis in Section \ref{section:results} highlights privacy-accuracy tradeoff. Where on this tradeoff spectrum we choose to lie depends on factors such as the motivation behind generating recourse in the first place, the context and end users behind the model, and the risk of an adversary taking hold of data and our trained model. Further research is needed to assess the human interpretability of the new recourses.

Figures \ref{fig:synth} and \ref{fig:hists} show that as dimension of synthetic data increases---especially as dimension passes the interpolation threshold (i.e., when the number of
training points equals the dimension: $d = n = 5000$)---MI attack success remains difficult to prevent, even under differentially private recourse generation methods. On high-dimensional datasets with too few training points, this finding highlights a continued privacy concern (even though such datasets are discouraged in the real world for accuracy and privacy purposes). Addressing this difficulty is worth further exploration.

There is practical value in exploring privacy risk mitigation in recourse-based membership inference attacks when the underlying model is a non-linear or non-inherently interpretable classifier, such as a neural network. \cite{abadi} propose a differentially private stochastic gradient descent (DP-SGD) mechanism by adding Gaussian distributed random noise to gradients during training. For neural network classifiers---and for general classifiers trained with gradient descent---we hypothesize that DP-SGD with carefully tuned hyperparameters can help us achieve recourse with lowered privacy risk.

\section{Conclusion}

We develop two methods to generate differentially private recourse, in order to protect against MI attacks leveraging explainability. We find particular success in preventing privacy leakage while maintaining model and recourse accuracy with \texttt{LR} (Laplace recourse), especially on \texttt{CFD} attacks, and especially when training dataset size is larger than dimension. 

Our work leaves behind remaining difficulties in this field. \texttt{CFD LRT} attacks remain more effective than \texttt{CFD} attacks. On synthetic datasets of high dimensionality, attacks remain hard to prevent, as distances are simply greater in high dimensions. While our \texttt{LR} method appears promising, research is needed to determine if the Laplace mechanism harms the human interpretability of the new recourses, in particular under high strength levels of differential privacy. Finally, while we considered a logistic regression classifier, this is an inherently interpretable model; it would be worthwhile investigating whether our results generalize to a blackbox neural network scenario.

\section{Acknowledgements}
The corresponding authors would like to thank Professor Hima Lakkaraju for motivating this work and for thoughtful feedback and discussions throughout the project. We also thank Postdoctoral Fellow Jiaqi Ma for their detailed guidance and feedback throughout the research process. We would like to also thank Martin Pawelczyk for his guidance on the implementation and evaluation portions of this work, and for clarifying to us ideas from his work (\citeauthor{recourseattack}).
Lastly, we thank the reviewers for their valuable comments and suggestions to improve our work. 

\newpage







\nocite{langley00}

\bibliography{paper}

\begin{thebibliography}{16}
\providecommand{\natexlab}[1]{#1}
\providecommand{\url}[1]{\texttt{#1}}
\expandafter\ifx\csname urlstyle\endcsname\relax
  \providecommand{\doi}[1]{doi: #1}\else
  \providecommand{\doi}{doi: \begingroup \urlstyle{rm}\Url}\fi

\bibitem[Abadi et~al.(2016)Abadi, Chu, Goodfellow, McMahan, Mironov, Talwar,
  and Zhang]{abadi}
Abadi, M., Chu, A., Goodfellow, I., McMahan, H.~B., Mironov, I., Talwar, K.,
  and Zhang, L.
\newblock Deep learning with differential privacy.
\newblock In \emph{Proceedings of the 2016 ACM SIGSAC conference on computer
  and communications security}, pp.\  308--318, 2016.

\bibitem[A{\"\i}vodji et~al.(2020)A{\"\i}vodji, Bolot, and
  Gambs]{aivodji2020model}
A{\"\i}vodji, U., Bolot, A., and Gambs, S.
\newblock Model extraction from counterfactual explanations.
\newblock \emph{arXiv preprint arXiv:2009.01884}, 2020.

\bibitem[Carlini et~al.(2021)Carlini, Chien, Nasr, Song, Terzis, and
  Tram{\`{e}}r]{carlini}
Carlini, N., Chien, S., Nasr, M., Song, S., Terzis, A., and Tram{\`{e}}r, F.
\newblock Membership inference attacks from first principles.
\newblock \emph{CoRR}, abs/2112.03570, 2021.
\newblock URL \url{https://arxiv.org/abs/2112.03570}.

\bibitem[Chaudhuri et~al.(2009)Chaudhuri, Monteleoni, and Sarwate]{dplogreg}
Chaudhuri, K., Monteleoni, C., and Sarwate, A.~D.
\newblock Differentially private support vector machines.
\newblock \emph{CoRR}, abs/0912.0071, 2009.
\newblock URL \url{http://arxiv.org/abs/0912.0071}.

\bibitem[Dua \& Graff(2017)Dua and Graff]{adult}
Dua, D. and Graff, C.
\newblock {UCI} machine learning repository, 2017.
\newblock URL \url{http://archive.ics.uci.edu/ml}.

\bibitem[Dwork \& Roth(2014)Dwork and Roth]{dwork}
Dwork, C. and Roth, A.
\newblock The algorithmic foundations of differential privacy.
\newblock \emph{Found. Trends Theor. Comput. Sci.}, 9\penalty0
  (3–4):\penalty0 211–407, aug 2014.
\newblock ISSN 1551-305X.
\newblock \doi{10.1561/0400000042}.
\newblock URL \url{https://doi.org/10.1561/0400000042}.

\bibitem[{FICO Community}()]{heloc}
{FICO Community}.
\newblock Explainable machine learning challenge.
\newblock URL
  \url{https://community.fico.com/s/explainable-machine-learning-challenge?tabset-158d9=3}.

\bibitem[Holohan et~al.(2019)Holohan, Braghin, Mac~Aonghusa, and
  Levacher]{diffprivlib}
Holohan, N., Braghin, S., Mac~Aonghusa, P., and Levacher, K.
\newblock Diffprivlib: the {IBM} differential privacy library.
\newblock \emph{ArXiv e-prints}, 1907.02444 [cs.CR], July 2019.

\bibitem[Jayaraman \& Evans(2019)Jayaraman and Evans]{evaluatingdpinpractice}
Jayaraman, B. and Evans, D.
\newblock Evaluating differentially private machine learning in practice, 2019.

\bibitem[Khodabakhsh et~al.(2019)Khodabakhsh, LeCun, Cortes, and Burges]{mnist}
Khodabakhsh, H., LeCun, Y., Cortes, C., and Burges, C.~J.
\newblock Mnist dataset, 2019.
\newblock URL \url{https://www.kaggle.com/datasets/hojjatk/mnist-dataset}.

\bibitem[Milli et~al.(2019)Milli, Schmidt, Dragan, and Hardt]{milli2019model}
Milli, S., Schmidt, L., Dragan, A.~D., and Hardt, M.
\newblock Model reconstruction from model explanations.
\newblock In \emph{Proceedings of the Conference on Fairness, Accountability,
  and Transparency}, pp.\  1--9, 2019.

\bibitem[Pawelczyk et~al.(2022)Pawelczyk, Lakkaraju, and Neel]{recourseattack}
Pawelczyk, M., Lakkaraju, H., and Neel, S.
\newblock On the privacy risks of algorithmic recourse.
\newblock \emph{arXiv}, abs/2211.05427, 2022.
\newblock URL \url{https://arxiv.org/abs/2211.05427}.

\bibitem[Shokri et~al.(2021)Shokri, Strobel, and Zick]{featureattack}
Shokri, R., Strobel, M., and Zick, Y.
\newblock On the privacy risks of model explanations.
\newblock \emph{arXiv}, abs/1907.00164, 2021.
\newblock URL \url{https://arxiv.org/abs/1907.00164}.

\bibitem[Wachter et~al.(2017)Wachter, Mittelstadt, and
  Russell]{wachter2017counterfactual}
Wachter, S., Mittelstadt, B., and Russell, C.
\newblock Counterfactual explanations without opening the black box: Automated
  decisions and the gdpr.
\newblock \emph{Harv. JL \& Tech.}, 31:\penalty0 841, 2017.

\bibitem[Weller(2019)]{transparency}
Weller, A.
\newblock Transparency: Motivations and challenges.
\newblock \emph{arXiv}, abs/1708.01870, 2019.
\newblock URL \url{https://arxiv.org/abs/1708.01870}.

\bibitem[Yeom et~al.(2017)Yeom, Fredrikson, and Jha]{yeom}
Yeom, S., Fredrikson, M., and Jha, S.
\newblock The unintended consequences of overfitting: Training data inference
  attacks.
\newblock \emph{CoRR}, abs/1709.01604, 2017.
\newblock URL \url{http://arxiv.org/abs/1709.01604}.

\end{thebibliography}
\bibliographystyle{icml2023}

\newpage
\appendix
\onecolumn
\section*{Appendix}


\subsection{\texttt{CFD LRT} Attack Detailed Formulation} \label{appendix:cfdlrt}

Section \ref{section:miattacks} introduces the \texttt{CFD LRT} attack that \cite{recourseattack} show successfully leak sensitive training data membership. To fully estimate the likelihood ratio, the adversary uses maximum likelihood estimation (MLE) methods to model the distributions of counterfactual distances (CFDs) when \textbf{1)} $x$ (the point in question) is in the training data, and when \textbf{2)} $x$ is in the test data. These building blocks comprise the numerator and denominator, respectively, of the likelihood ratio $\Lambda$. 

\cite{recourseattack} models the distributions of CFDs as log-normal (parameterized by mean and standard deviation), meaning MLE estimates are of $(\mu_{in}, \sigma_{in})$ and $(\mu_{out}, \sigma_{out}).$ In the full estimation process, the adversary — who has access to training data distribution $\mathcal{D}$ — can estimate all four parameters by training shadow models with and without point $x$, and then computing the resulting CFDs.

However, this process would entail sampling an adversary training set and training a shadow model separately for each data entry $x$ that we perform the attack on. This is computationally intractable in practice. To work around this, Pawelczyk et al. show and implement a \textit{one-sided} version of the LRT, where we only estimate $\mu_{out}, \sigma_{out}$, and the attack predicts \texttt{MEMBER} if $c(x,x')$'s likelihood ratio is sufficiently low under such parameters. Conveniently, since $\mu_{out}, \sigma_{out}$ do not depend on $x$, we need only train shadow models once.

Algorithm 1 shows a detailed formulation of the proposed one-sided \texttt{CFD LRT} attack, tailored to the linear classifier case.

\begin{algorithm} 
\label{algorithm:cfdlrt}
\caption{CFD-based Likelihood Ratio Test (\texttt{CFD LRT}) for Linear Classifier}
\begin{algorithmic}
\STATE \textbf{Inputs:} $x$: point in question. $t_0 = c(x, x')$: $x$'s CFD in the model trained on owner training data. $\mathcal{D}$: training data distribution. $\alpha$: \texttt{FPR}. $N$: number of shadow models.
\STATE estimatedCFD = []
\STATE \textbf{Compute:} $t_0 = c(x, x')$
\FOR {$i \gets 1,\dots,N$}
    \STATE Sample $\mathcal{D}_t^{(i)} \sim \mathcal{D}$ 
    \hfill\COMMENT{Adversary's training set for training shadow model $i$.}
    \STATE $f_{\theta^{(i)}} = \texttt{TrainShadowClassifier}(\mathcal{D}_t^{(i)})$ \hfill\COMMENT{$i$th shadow model.}
    \STATE $c(x, x'^{(i)}) = \texttt{GetCFD}(x, f_{\theta^{(i)}})$
    \STATE estimatedCFD $\leftarrow c(x, x'^{(i)})$
\ENDFOR
\STATE $\hat{\mu}_{\text{out, MLE}} = \frac{1}{N}\sum_{i=1}^{N} \left(\log c(x, \mathbf{x'}^{(i)})\right)$
\STATE $\hat{\sigma}^2_{\text{out, MLE}} = \frac{1}{N}\sum_{i=1}^{N} \left(\hat{\mu}_{\text{out, MLE}} - \big(\log c(x, \mathbf{x'}^{(i)})\big)\right)^2$
\COMMENT{$z_{1-\alpha}$ is the $1-\alpha$ quantile of $Z \sim \mathrm{LogNormal}(\hat{\mu}_{\text{out, MLE}}, \hat{\sigma}^2_{\text{out, MLE}}).$}
\IF {$t_0 > z_{1-\alpha}$} 
    \STATE \textbf{Output:} \texttt{NON-MEMBER}
\ELSE 
    \STATE \textbf{Output:} \texttt{MEMBER}
\ENDIF
\end{algorithmic}
\end{algorithm}

\subsection{Experimental Results on Adult Dataset} \label{section:adultappendix}

\begin{figure}[h!]
    \centering
    \includegraphics[width=0.5\linewidth]{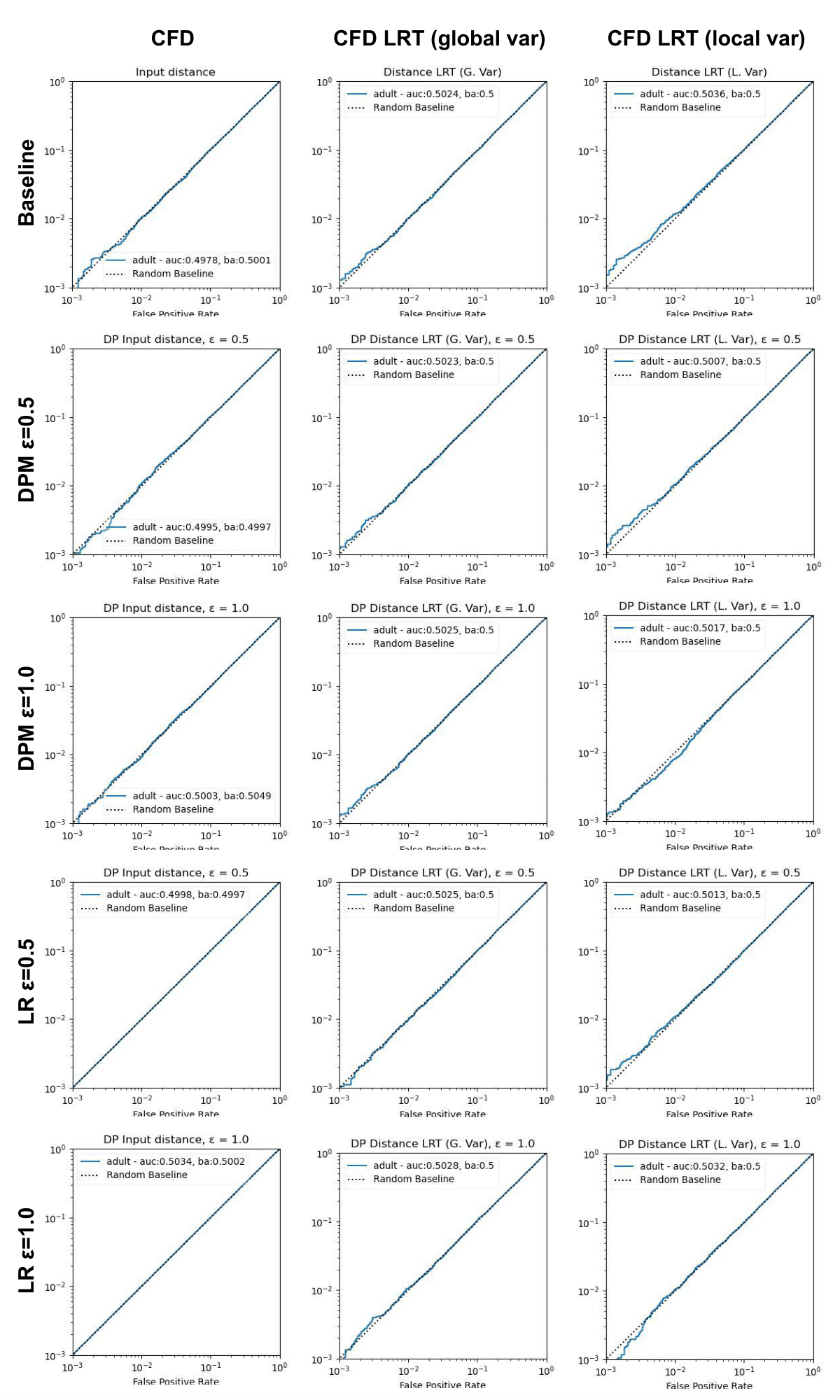}
    \caption{Log-scaled \texttt{ROC} curves (\texttt{TPR} v. \texttt{FPR}), \texttt{AUC}, and \texttt{BA} for all attacks on all settings using Adult dataset. \texttt{ROC} curves are near-random and \texttt{AUC} is around 0.5 even on baseline, signifying a lack of success of the attacks and therefore a negligible potential for our DP methods to help.}
    \label{fig:adult}
\end{figure}

See Fig. \ref{fig:adult}.

\end{document}